\documentclass[conference]{IEEEtran}
\usepackage{cite}
\usepackage{amsmath,amssymb,amsfonts}
\usepackage{algorithmic}
\usepackage{graphicx}
\usepackage{textcomp}

\usepackage{silence}
\usepackage[dvipsnames]{xcolor} % Extended color names (dvipsnames loaded via \PassOptionsToPackage in main.tex)
\usepackage{soul}               % Text highlighting
\definecolor{light_gray}{rgb}{0.7, 0.7, 0.7}
\definecolor{gray}{gray}{0.85}
\definecolor{cusyellow}{rgb}{1, 0.706, 0}
\definecolor{cusblue}{rgb}{0, 0.651, 0.929}

\usepackage{pifont}
\newcommand{\cmark}{\textcolor{ForestGreen}{\ding{51}}}
\newcommand{\xmark}{\textcolor{red}{\ding{55}}}

\usepackage{amsmath, amssymb, amsfonts, bm}

\usepackage{subcaption, booktabs, multirow, array, colortbl, bbding, longtable}

\usepackage{cleveref}
\usepackage{subcaption}   % Sub-figures
\usepackage{booktabs}     % Professional table lines
\usepackage{multirow}     % Multi-row cells in tables
\usepackage{array}        % Advanced table column formatting
\usepackage{colortbl}     % Color in tables
\usepackage{bbding}       % Extra symbols (e.g., \XSolidBrush, \Checkmark)
\usepackage{longtable}    % Multi-page tables
\usepackage{xspace}

\usepackage{makecell}
\newcommand{\ours}{POCI-Diff}

\def\BibTeX{{\rm B\kern-.05em{\sc i\kern-.025em b}\kern-.08em
    T\kern-.1667em\lower.7ex\hbox{E}\kern-.125emX}}
\begin{document}

\title{\ours: 3D-Layout Guided Diffusion for Controllable Synthetic Surveillance Data Generation\\
% {\footnotesize \textsuperscript{*}Note: Sub-titles are not captured for https://ieeexplore.ieee.org  and
}

% \author{Andrea Rigo\inst{1} \and
% Luca Stornaiuolo\inst{2} \and
% Weijie Wang\inst{1,4} \and
% Mauro Martino\inst{3} \and
% Bruno Lepri\inst{4} \and
% Nicu Sebe\inst{1}}
% % \author{First Author\inst{1}\orcidID{0000-1111-2222-3333} \and
% % Second Author\inst{2,3}\orcidID{1111-2222-3333-4444} \and
% % Third Author\inst{3}\orcidID{2222--3333-4444-5555}}

% \authorrunning{A. Rigo et al.}

% \institute{DISI, University of Trento, Trento, Italy \and
% Toretei S.r.l., Roma, Italy \and
% Visual AI Lab, MIT-IBM Watson AI Lab, Cambridge, Massachusetts, U.S.A. \and
% Fondazione Bruno Kessler, Trento, Italy
% }

\author{\IEEEauthorblockN{1\textsuperscript{st} Andrea Rigo}
\IEEEauthorblockA{\textit{University of Trento}\\
Trento, Italy \\
andrea.rigo@unitn.it}
\and
\IEEEauthorblockN{2\textsuperscript{nd} Luca Stornaiuolo}
\IEEEauthorblockA{\textit{Toretei S.r.l.}\\
Roma, Italy \\
luca.stornaiuolo@polimi.it}
\and
\IEEEauthorblockN{3\textsuperscript{rd} Weijie Wang}
\IEEEauthorblockA{\textit{University of Trento}\\
Trento, Italy \\
weijie.wang@unitn.it}
%\and
\and
\and[\newline]
\and
\IEEEauthorblockN{4\textsuperscript{th} Mauro Martino}
\IEEEauthorblockA{\textit{MIT-IBM Watson AI Lab}\\
Cambridge, MA, USA \\
mmartino@us.ibm.com}
\and
\IEEEauthorblockN{5\textsuperscript{th} Bruno Lepri}
\IEEEauthorblockA{\textit{Fondazione Bruno Kessler}\\
Trento, Italy \\
lepri@fbk.eu}
\and
\IEEEauthorblockN{6\textsuperscript{th} Nicu Sebe}
\IEEEauthorblockA{\textit{University of Trento}\\
Trento, Italy \\
niculae.sebe@unitn.it}
}

\maketitle

\begin{abstract}
% === ORIGINAL ABSTRACT ===
% We propose a diffusion-based approach for Text-to-Image (T2I) generation with consistent and interactive 3D layout control and editing. While prior methods improve spatial adherence using 2D cues or iterative copy-warp-paste strategies, they often distort object geometry and fail to preserve consistency across edits.
% To address these limitations, we introduce a framework for \emph{P}ositioning \emph{O}bjects \emph{C}onsistently and \emph{I}nteractively (POCI-Diff), a novel formulation for jointly enforcing 3D geometric constraints and instance-level semantic binding within a unified diffusion process.
% Our method enables explicit per-object semantic control by binding individual text descriptions to specific 3D bounding boxes through Blended Latent Diffusion, allowing one-shot synthesis of complex multi-object scenes. We further propose a warping-free generative editing pipeline that supports object insertion, removal, and transformation via regeneration rather than pixel deformation. To preserve object identity and consistency across edits, we condition the diffusion process on reference images using IP-Adapter, enabling coherent object appearance throughout interactive 3D editing while maintaining global scene coherence. Experimental results demonstrate that POCI-Diff produces high-quality images consistent with the specified 3D layouts and edits, outperforming state-of-the-art methods in both visual fidelity and layout adherence while eliminating warping-induced geometric artifacts.
% === END ORIGINAL ABSTRACT ===
Training robust visual surveillance models requires large-scale datasets with precise spatial annotations,
yet collecting real surveillance data is costly, privacy-sensitive, and often legally constrained.
Synthetic data generation offers a compelling alternative, but existing methods lack fine-grained 3D
control over object placement and appearance, limiting geometric consistency across camera viewpoints.
We introduce \ours{} (\emph{P}ositioning \emph{O}bjects \emph{C}onsistently and \emph{I}nteractively),
a framework that generates annotated synthetic scenes from explicit 3D bounding-box layouts
with per-object semantic control.
By integrating Blended Latent Diffusion with depth-conditioned ControlNet, \ours{} synthesises
complex multi-object scenes in a single forward pass, binding individual text descriptions to specific
3D locations.
We further propose a warping-free editing pipeline supporting object insertion, removal, and
transformation via regeneration, enabling efficient scene variation for data augmentation.
Object identity across edits is preserved by conditioning on reference images via IP-Adapter,
ensuring appearance consistency throughout interactive scene manipulation.
Experiments show that \ours{} outperforms state-of-the-art 3D layout-guided generation methods
in visual fidelity and layout adherence, while eliminating warping-induced geometric artifacts.

\end{abstract}


\section{Introduction}
\label{sec:intro}

Training robust surveillance models for multi-object tracking (MOT), re-identification, and anomaly detection demands large annotated datasets, yet real footage is costly to collect and subject to strict privacy regulations~\cite{delussu2024synthetic_surv_review}.
% Synthetic data generation offers a scalable alternative~\cite{farley2023diffusion_ped_det,seib2024synth_ped_gan,niu2024diffusion_reid}, but existing methods lack fine-grained 3D control over object placement and appearance, a critical limitation for fixed-viewpoint cameras with known geometry.
Synthetic data generation offers a scalable alternative~\cite{farley2023diffusion_ped_det,niu2024diffusion_reid}, but existing methods lack fine-grained 3D control over object placement and appearance, a critical limitation for fixed-viewpoint cameras with known geometry.
Since surveillance cameras are fixed, training data must match their viewpoint and allow controlled variation in crowd density, occlusion patterns, and object configurations, a capability that existing generators do not support.

Recent T2I diffusion models deliver high visual realism, but layout-control methods remain limited to 2D cues, keypoints, bounding boxes, or segmentation masks~\cite{li2023gligen,xie2023boxdiff,ca_guidance}, which cannot enforce geometrically consistent object placement in 3D space.
LooseControl~\cite{bhat2023loosecontrol} takes a step forward by conditioning a ControlNet~\cite{controlnet} on depth maps rendered from 3D bounding boxes, but uses a single global text prompt and therefore cannot associate object-specific descriptions with individual locations.
Build-a-Scene (BAS)~\cite{eldesokey2024buildasceneinteractive3dlayout} adds per-object semantic control by inserting objects one at a time into a background scene using 3D layout guidance.
To edit an object's position, BAS crops it from the image, warps it to the target 3D location, and lightly harmonises it via diffusion, a copy-warp-paste strategy that distorts object geometry under large spatial transformations and produces blending inconsistencies.
Sequential insertion also makes the pipeline computationally inefficient and prone to error accumulation.

\begin{figure}[t]
    \centering
    \includegraphics[width=0.9\linewidth]{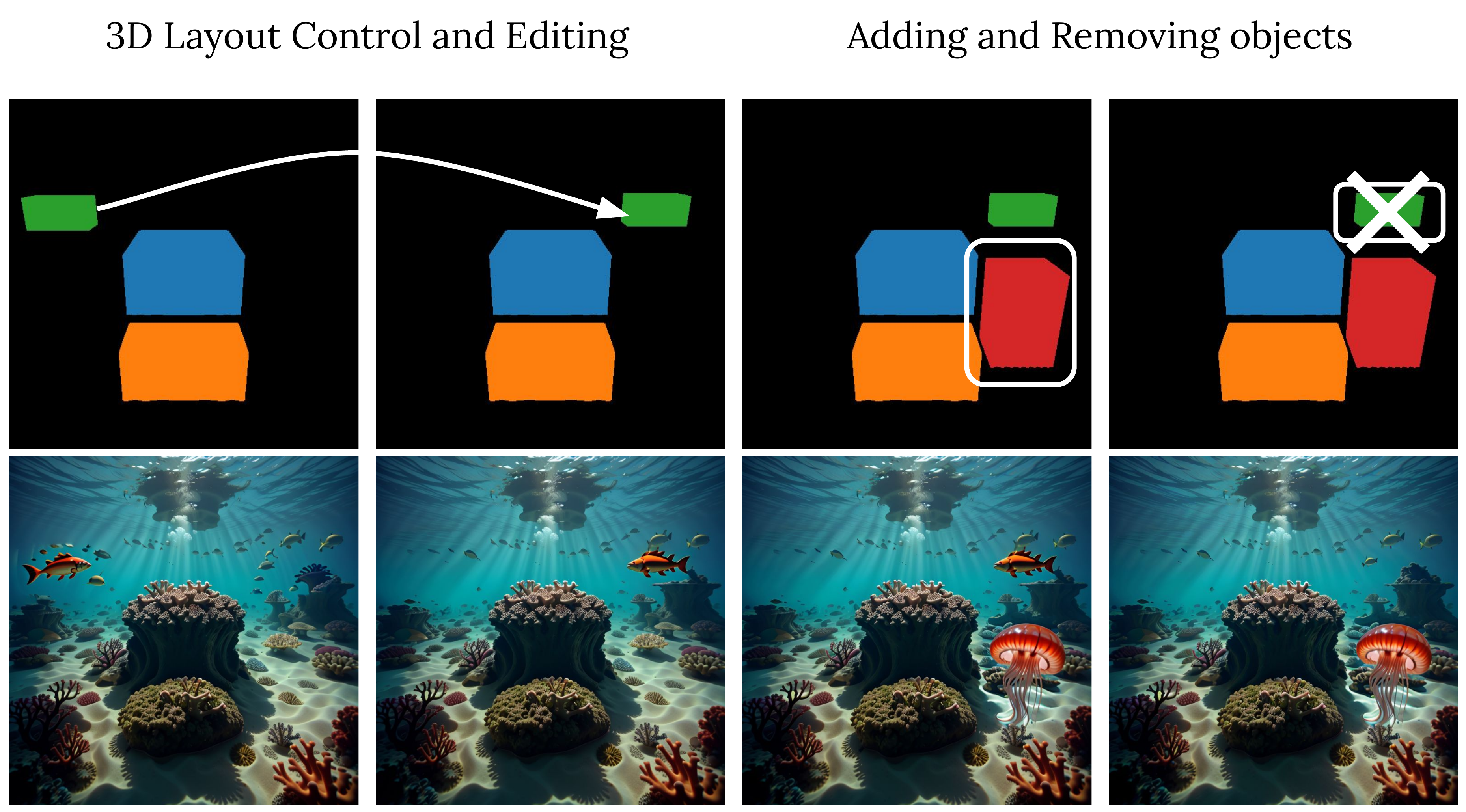}
    % === ORIGINAL CAPTION ===
    % \caption{\ours{} introduces an interactive diffusion-based system for image generation from 3D layouts, enabling precise control over object type, position, scale, and orientation, as well as camera movement, while supporting layout-driven image editing with seamless integration of new or modified objects.}
    % === END ORIGINAL CAPTION ===
    \caption{\ours{} synthesises photorealistic scenes from 3D bounding-box layouts with per-object control over type, position, scale, and orientation. Its warping-free editing pipeline supports interactive object insertion, removal, and transformation for efficient data augmentation.}
    \label{fig:teaser}
    % \vspace{-0.4cm}
\end{figure}

\begin{table}[t]
\centering
\tabcolsep 3pt
\resizebox{\linewidth}{!}{
\begin{tabular}{lccccc}
\hline
\multicolumn{1}{c}{\multirow{2}{*}{Method}} &
Per-object
 & One-shot 3D &
Warping-free   &
Computational \\
\multicolumn{1}{c}{}  &  Semantic Control  & Scene Generation & Editing &   Efficiency  \\ \hline
LooseControl~\cite{bhat2023loosecontrol}    & \xmark  & \cmark & \cmark    & \cmark  \\
Build-A-Scene~\cite{eldesokey2024buildasceneinteractive3dlayout}       & \cmark  & \xmark (iterative only)  & \xmark (warp distortion)         & \xmark    \\
\ours{} (Ours)       & \cmark   & \cmark   & \cmark    & \cmark         \\ \hline
\end{tabular}}
\caption{Comparison of key capabilities across methods.
POCI-Diff is the only method that simultaneously supports one-shot scene generation from 3D layouts, per-object semantic control, and warping-free editing with high efficiency.
}
\label{tab:intro:diff}
\vspace{-0.5cm}
\end{table}

We propose \ours{} (\cref{fig:teaser}), a unified framework that addresses all three limitations.
By integrating Blended Latent Diffusion~\cite{avrahami2023blended} with a fully fine-tuned depth-conditioned ControlNet~\cite{controlnet}, \ours{} synthesises complex multi-object scenes in a single forward pass, binding individual text prompts to specific 3D bounding boxes.
Unlike LooseControl and BAS, which use lightweight LoRA-based ControlNet adaptation and exhibit box-like geometric artifacts, \ours{} fine-tunes the ControlNet on high-resolution synthetic images from Flux.1~\cite{flux2024}, enabling clean interpretation of volumetric 3D constraints.
For interactive editing, \ours{} takes a warping-free approach: instead of cropping and geometrically deforming pixels as in BAS, it inpaints the source region and regenerates the object at its new 3D location directly from the updated layout.
This avoids the geometric distortions introduced by warping-based methods.
To preserve object appearance throughout manipulation, we condition the diffusion process on a reference crop of the object via IP-Adapter~\cite{ye2023ip}, ensuring consistent identity across insertions, removals, and spatial transformations.
\cref{tab:intro:diff} summarises our advantages over prior work.

Our contributions are:
\begin{itemize}
    \item \ours{}, a depth-guided diffusion framework for one-shot scene generation from 3D bounding-box layouts with fine-grained per-object semantic control, directly applicable to generating diverse, privacy-compliant surveillance training data.
    \item A warping-free editing pipeline combining Blended Latent Diffusion with a two-stage IP-Adapter conditioning strategy, supporting interactive object insertion, removal, and transformation while preserving object identity and global scene coherence.
    \item Quantitative and qualitative evaluations, including a 50-participant user study, demonstrating state-of-the-art performance in 3D layout controllability, text-image alignment, and visual fidelity.
\end{itemize}

\section{Related Works}
\label{sec:related-works}

\subsection{Synthetic Data for Visual Surveillance}
Synthetic data has long been used as a substitute for real surveillance footage, which is expensive to collect and constrained by privacy regulations~\cite{delussu2024synthetic_surv_review}.
% Early synthetic approaches use 3D graphics pipelines for precise annotations~\cite{hattori2015learning_scene_specific_ped}.
% Recent methods adopt generative models to improve photorealism: GAN-based pipelines synthesise pedestrian scenes from semantic maps~\cite{seib2024synth_ped_gan}, diffusion augmentation closes the sim-to-real gap for detection~\cite{farley2023diffusion_ped_det}, and identity-driven diffusion addresses re-identification data scarcity~\cite{niu2024diffusion_reid,dai2025dive_vi_reid}.
Recent methods adopt generative models to improve photorealism: diffusion augmentation closes the sim-to-real gap for detection~\cite{farley2023diffusion_ped_det}, and identity-driven diffusion addresses re-identification data scarcity~\cite{niu2024diffusion_reid,dai2025dive_vi_reid}.
Despite this progress, none of these methods offer explicit 3D control over object placement from a known camera viewpoint.
\ours{} fills this gap with 3D layout-guided generation and interactive editing.

\subsection{3D Layout Control and Scene Editing}
% Most layout-conditioned diffusion methods rely on 2D spatial cues, bounding boxes, keypoints, or segmentation masks, either via fine-tuned attention mechanisms~\cite{li2023gligen,zheng2023layoutdiffusion} or training-free cross-attention guidance~\cite{xie2023boxdiff,ca_guidance}.
Most layout-conditioned diffusion methods rely on 2D spatial cues, bounding boxes, keypoints, or segmentation masks, either via fine-tuned attention mechanisms~\cite{li2023gligen} or training-free cross-attention guidance~\cite{xie2023boxdiff,ca_guidance}.
These 2D approaches cannot enforce geometrically consistent object placement in 3D scenes.
LooseControl~\cite{bhat2023loosecontrol} addresses this by rendering depth maps from 3D bounding boxes to condition a ControlNet~\cite{controlnet}, improving geometric coherence but relying on a single global text prompt with no per-object semantic control.
Build-a-Scene (BAS)~\cite{eldesokey2024buildasceneinteractive3dlayout} adds per-object semantics through an iterative pipeline, but its copy-warp-paste editing strategy introduces geometric distortions, and sequential object insertion is computationally inefficient.
% A parallel line of research targets 3D-consistent multi-view synthesis via NeRF or mesh lifting~\cite{Hollein2024ViewDiff3I,UrbanGen2025}, but these methods do not accept 3D bounding-box layouts and target multi-view coherence rather than interactive single-image layout control.
A parallel line of research targets 3D-consistent multi-view synthesis via NeRF or mesh lifting~\cite{UrbanGen2025}, but these methods do not accept 3D bounding-box layouts and target multi-view coherence rather than interactive single-image layout control.

% For appearance consistency across edits, adapter-based methods~\cite{ye2023ip,ma2024subject} condition the diffusion model on reference images without full fine-tuning.
For appearance consistency across edits, adapter-based methods~\cite{ye2023ip} condition the diffusion model on reference images without full fine-tuning.
\ours{} combines 3D layout control, per-object semantic binding, one-shot scene generation, and reference-conditioned identity preservation into a unified warping-free framework.

\section{Method}
\label{sec:method}

\begin{figure}[t]
    \centering
    \includegraphics[width=1\linewidth]{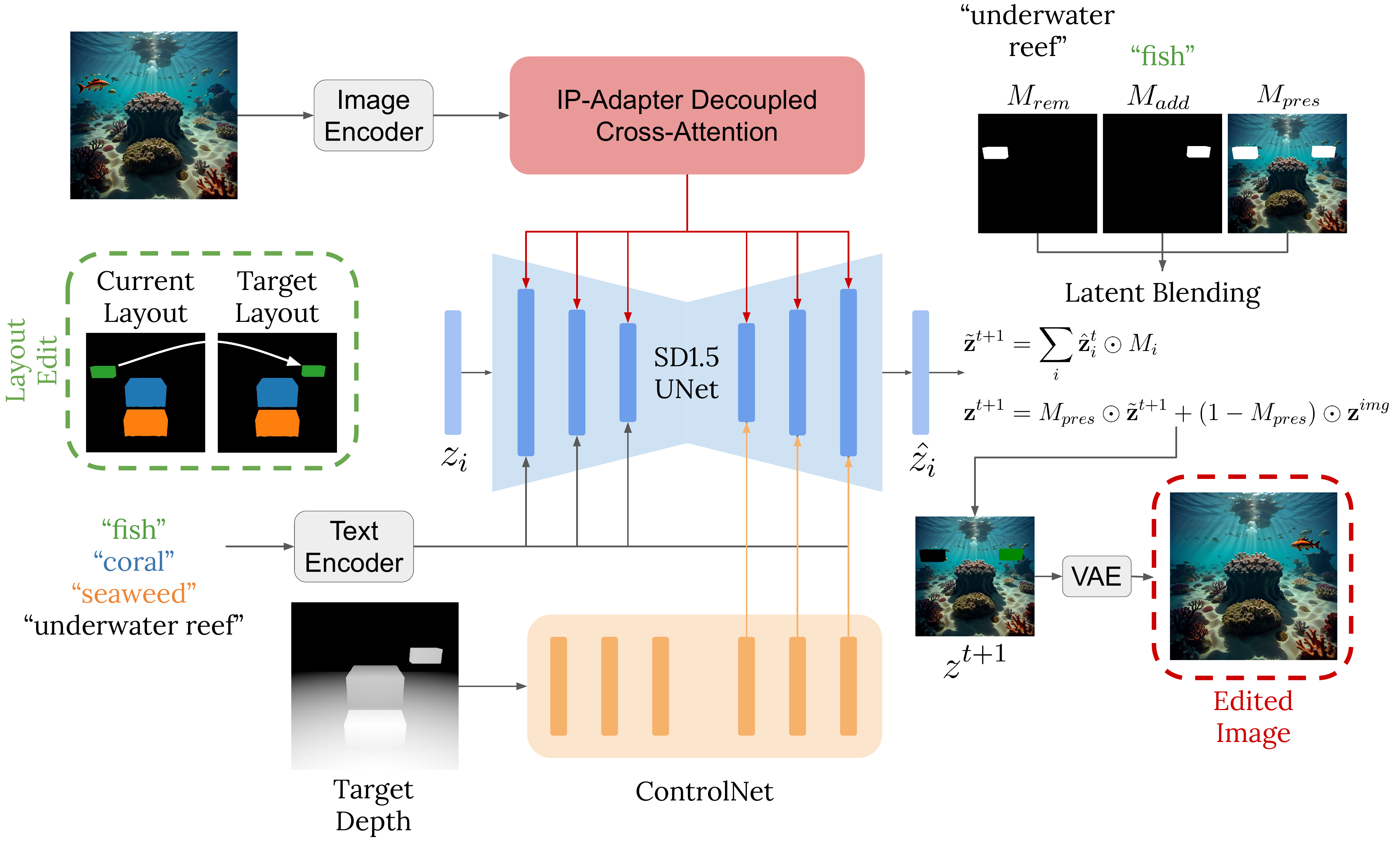}
    \caption{Overview of our proposed pipeline. \ours{} enables one-shot 3D layout–guided scene generation and warping-free object translation, preserving object identity while maintaining global scene harmonization.}
    \label{fig:pipeline}
    \vspace{-0.4cm}
\end{figure}

% (Original)
% The objective of 3D-layout guided text-to-image (T2I) diffusion is to synthesize images that faithfully adhere to spatial layouts and textual descriptions. Formally, given a layout defined by a set of 3D bounding boxes $B = \{B_1, B_2, \dots, B_n\}$ and their corresponding text prompts $P = \{P_1, P_2, \dots, P_n\}$, we aim to achieve two goals: 1) accurately generate the scene and requested objects within the spatial constraints of the 3D bounding boxes; and 2) facilitate comprehensive interactive scene manipulation. Specifically, we propose \ours{}, a pipeline that supports object insertion, removal, replacement, and spatial transformations (translation, rotation and scaling), as well as free viewpoint control, all while maintaining global scene consistency and high visual fidelity.
% These operations can be combined in any order, enabling the construction of complex, multi-step editing workflows.
% A detailed illustration of our proposed pipeline is shown in \cref{fig:pipeline}.

The goal of 3D-layout guided text-to-image (T2I) diffusion is to synthesize images that follow a 3D layout and textual descriptions. Given 3D bounding boxes $B = \{B_1, B_2, \dots, B_n\}$ and corresponding prompts $P = \{P_1, P_2, \dots, P_n\}$, we target faithful layout adherence and interactive, composable scene edits. We propose \ours{}, which supports object insertion, removal, and replacement; spatial transformations (translation, rotation, scaling); and free viewpoint control, while maintaining global scene consistency and high visual fidelity (\cref{fig:pipeline}).

\subsection{3D Layout Control}
\label{sec:pipe-base}
% Our goal is to synthesize all objects and the background within a single, unified diffusion process, rather than iteratively inserting objects one by one as in prior work.
% To this end, we jointly denoise all scene elements and merge them at every step, resulting in a single coherent image.
% To this end, we formulate scene generation as a set of parallel diffusion paths that are jointly denoised and merged at every step, resulting in a single coherent image.

To condition diffusion models on 3D layouts, we fine-tune a depth-conditioned ControlNet~\cite{controlnet} on 3D layout depth maps, enabling the model to interpret 3D bounding boxes as volumetric priors and suppress geometric distortions.
Although this approach guides generation with the layout and a global text prompt, it lacks support for per-object descriptions. Consequently, although the scene geometry is controlled, the semantic identity of the object within each box cannot be explicitly defined.
% Although this approach ensures generation is guided by the layout and a global text prompt, it lacks support for bounding-box-specific descriptions. Consequently, although the scene geometry is controlled, the semantic identity of the object within each box cannot be explicitly defined.
% Build-a-Scene \cite{eldesokey2024buildasceneinteractive3dlayout} (BAS) uses LooseControl as backbone for layout control, and allows bounding-box-specific description by iteratively adding each object to the scene.
%, starting from the empty background.
We address this limitation by incorporating Blended Latent Diffusion \cite{avrahami2023blended} (BLD) to assign specific text prompts to individual bounding boxes.
This allows our approach, which we name \ours, to efficiently generate the whole scene from the input 3D layout in one go, in contrast with the inefficient iterative approach of BAS.
In addition to the depth-conditioned ControlNet, for each bounding box, we generate a mask $M_i$ by projecting the box onto the image plane, excluding occluded regions.
% In addition to the depth-conditioned ControlNet, for each bounding box, we generate a mask $M_i$ by projecting the box onto the image plane. We procedurally refine these masks to exclude occluded regions, ensuring the pipeline respects proper depth ordering.
We define the background mask $M_{bg}$ as the inverse of the union of all object masks. The scene is thus represented by the set of masks $M = \{M_1, M_2, ..., M_{n}, M_{bg}\}$, along with their corresponding prompts. 
Inspired by BLD, we execute parallel independent diffusion paths for each object and the background. At each denoising step, we merge the resulting latents:
% Although our procedural refinement minimizes overlaps, we employ a pixel-wise normalization strategy similar to MultiDiffusion \cite{bartal2023multidiffusionfusingdiffusionpaths} to ensure robustness against refinement errors:
% Inspired by BLD, we execute parallel independent diffusion paths for each object and the background. At each denoising step, we merge the resulting latents. Although our procedural refinement minimizes overlaps, we employ a pixel-wise normalization strategy similar to MultiDiffusion \cite{bartal2023multidiffusionfusingdiffusionpaths} to ensure robustness against refinement errors:

\begin{equation}
\label{eq:blend}
    \mathbf{z}^{t+1} = \sum_{i} \hat{\mathbf{z}^t_i} \odot M_i
\end{equation}
% \begin{equation}
% \label{eq:blend}
%     \mathbf{z}^{t+1} = \frac{\sum_{i} \hat{\mathbf{z}^t_i} \odot M_i}{\epsilon + \sum_{i} M_i}
% \end{equation}
% 
where $\hat{\mathbf{z}^t_i}$ represents the latent prediction for the $i$-th input (object or background) at step $t$, $M_i$ denotes the corresponding spatial mask, and $\odot$ indicates the Hadamard product.
% The denominator performs element-wise normalization using the sum of mask weights (plus $\epsilon$ for numerical stability), ensuring the fused representation remains within the model’s valid distribution even if masks overlap.
The composed latent $\mathbf{z}^{t+1}$ serves as the input for the subsequent diffusion step for all objects.
This procedure conditions the model on the 3D layout with per-box text descriptions, while the shared denoising steps harmonize the objects, resulting in a cohesive scene.

In the editing pipeline detailed below, we condition the diffusion process on the source image using IP-Adapter \cite{ye2023ip}, which facilitates object repositioning while preserving stylistic consistency. IP-Adapter employs a decoupled cross-attention mechanism to inject image features separately from text features. Empirically, we observed that a direct application of IP-Adapter failed to maintain sufficient consistency between the new and original objects. However, we found that consistency is preserved when both objects are generated using the same reference signal. Consequently, our pipeline adopts a two-stage approach: first, we generate a reference image from the input control layout; second, we regenerate the scene with the object diffusion processes conditioned on this reference via IP-Adapter.
This allows us to condition subsequent editing steps on the reference image originally used to generate the object and thus edit it while preserving its appearance.
In practice, we found that the resulting output image can replace the reference image without compromising edit consistency, thereby eliminating the need to store the reference.
% In practice, we observed that the resulting output image can effectively replace the initial reference image for subsequent editing tasks, eliminating the need to store the reference image.

\subsection{Adding, Removing and Replacing Objects}
\label{sec:pipe-add-remove}
To enable interactive editing, we allow users to add, remove, or replace objects in the generated scene. We achieve this by re-running the pipeline described in \cref{sec:pipe-base}, using the updated 3D layout to condition the ControlNet. To integrate changes seamlessly, we employ BLD as an inpainting mechanism.
Specifically, we execute the diffusion process within the object mask $M_{obj}$, blending it at each time step with the existing image content defined by the inverse mask as follows:

\begin{equation}
    \label{eq:inpainting-mask}
    \mathbf{z}^{t+1} = M_{obj} \hat{\mathbf{z}^{t}} + (1-M_{obj}) \mathbf{z}^{img}
\end{equation}
where $\mathbf{z}^{img}$ is the image latent.

The text conditioning depends on the operation: for adding or replacing objects, the diffusion is guided by the new object's text description; for object removal, it is conditioned on the layout's global background prompt.

\subsection{Consistent 3D Editing}
\label{sec:3d-editing}
% Existing layout control approaches, including LooseControl, struggle to preserve object identity under geometric transformations such as scaling and translation.
% BAS attempts to address this by segmenting the object via Segment Anything (SAM) \cite{sam}, and warping it to a new location.
% The warped object is then pasted onto a clean background image saved prior to the object's original insertion, followed by DDIM inversion \cite{ddim}, and blending via BLD for a few diffusion steps to harmonize the result.
% However, as shown on the right of \cref{fig:edit-quals}, this "copy-warp-paste" workflow often results in significant geometric distortion and poor harmonization. Furthermore, the reliance on a pre-existing background image without the object limits its applicability in complex editing scenarios.

We propose a streamlined pipeline that moves an object by simultaneously regenerating it in the target location and erasing it from the source location, avoiding warping artifacts entirely (\cref{fig:pipeline}).
% Given a layout update, we define three spatial masks based on projections to the image plane:
Given a layout update, we define three spatial masks: an addition mask $M_{add}$ corresponding to the new bounding box; a removal mask $M_{rem}$ corresponding to the object's original bounding box; and a preservation mask $M_{pres}$, defined as the union of the other two regions.

We run the pipeline as in \cref{sec:pipe-base}, but guide $M_{add}$ with the object’s prompt to generate the object, and $M_{rem}$ with the background prompt to inpaint the area left behind.
% We execute the diffusion pipeline described in \cref{sec:pipe-base}, but with specific conditioning for each region. The diffusion path for $M_{add}$ is conditioned on the object's text description to generate the object, while the path for $M_{rem}$ is conditioned on the global background prompt to inpaint the void left by the object.
At each step, we harmonize the active regions $M = \{M_{add}, M_{rem}\}$ using the blending in \cref{eq:blend}, then we apply the inpainting operation defined in \cref{eq:inpainting-mask} using $M_{pres}$ and the original image latent $\mathbf{z}^{img}$ to preserve the rest of the image.
% First we harmonize the active regions $M = \{M_{add}, M_{rem}\}$ at each step using the blending in \cref{eq:blend}, then we apply the inpainting operation defined in \cref{eq:inpainting-mask} using $M_{pres}$ and the original image latent $\mathbf{z}^{img}$ to strictly preserve unaltered parts of the image:

% \begin{equation}
% \begin{aligned}
% \tilde{\mathbf{z}}^{t+1}
% &=
% \sum_i \hat{\mathbf{z}}^t_i \odot M_i
% \qquad\qquad
% \mathbf{z}^{t+1}
% =
% M_{pres} \odot \tilde{\mathbf{z}}^{t+1}
% +
% (1 - M_{pres}) \odot \mathbf{z}^{img}
% \end{aligned}
% \end{equation}

% Crucially, to ensure the object generated at the new coordinates retains the visual identity of the original, we condition the diffusion process on the source image via IP-Adapter \cite{ye2023ip}. As explained in \cref{sec:pipe-base}, we achieve consistency by first synthesizing an intermediate reference image, and then generating the previous stage's output conditioned on this reference. This ensures that by conditioning the editing process on this same reference, the relocated object remains consistent with the original. In practice, however, we find that explicitly retaining this intermediate reference is unnecessary. We can effectively discard the reference image and condition the IP-Adapter directly on the previous stage's output, obtaining an equivalent level of consistency in the edited image.
Crucially, to ensure the edited object retains the visual identity of the original, we condition the diffusion process on the input image via IP-Adapter \cite{ye2023ip}, as explained in \cref{sec:pipe-base}.
% 
% This process allows our model to synthesize the object in the new location by referencing the style and semantics of the original object, while the ControlNet ensures adherence to the new 3D layout.
% The resulting pipeline achieves consistent 3D translation by regenerating the object with high stylistic fidelity and seamless background inpainting, ensuring global scene harmonization without the artifacts associated with pixel-space warping.
This enables object synthesis at the new location while preserving style and semantics, with ControlNet enforcing the target 3D layout. The resulting pipeline achieves consistent 3D translation with high stylistic fidelity and seamless background inpainting, avoiding warping artifacts.
% Additionally, by omitting the removal mask and resizing the object bounding box, this pipeline can scale the object up and down while keeping its style consistent.

\section{Experiments}
\label{sec:experiments}

\begin{table}[t]
\centering
\scriptsize
\setlength{\tabcolsep}{3pt}
\begin{tabular}{@{}c|lcccc@{}}
\toprule
 &  & \textbf{\ours{}} & BAS & LooseCtrl & LayoutG. \\ \midrule
\multirow{3}{*}{\rotatebox{90}{\makecell{Layout\\Control}}}
 & $\text{CLIP}_{T2I}$ \textuparrow & \textbf{0.333} & 0.321 & 0.302 & 0.323 \\
 & OA \textuparrow & \textbf{78.3} & 53.5 & 24.3 & 48.2 \\
 & mIOU \textuparrow & 0.728 & \textbf{0.772} & 0.633 & 0.425 \\
 \addlinespace[1.5pt]
 \midrule
\multirow{10}{*}{\rotatebox{90}{Consistency}}
 & Crop $\text{CLIP}_{I2I}$ \textuparrow & 0.919 & \textbf{0.924} & \textbf{0.924} & 0.838 \\
 & Crop PSNR \textuparrow & 29.10 & \textbf{29.3} & 29.12 & 28.35 \\
 & $\text{CLIP}_{I2I}$ \textuparrow & \textbf{97.068} & 96.398 & - & - \\
 & DINOv2 \textuparrow & \textbf{0.919} & 0.873 & - & - \\
 & LPIPS \textdownarrow & \textbf{0.115} & 0.154 & - & - \\
 & DISTS \textdownarrow & \textbf{0.081} & 0.118 & - & - \\
 & ARNIQA \textuparrow & \textbf{0.673} & 0.670 & - & - \\
 & ImageReward \textuparrow & \textbf{1.281} & 0.863 & - & - \\
 & CLIP-IQA \textuparrow & \textbf{0.884} & 0.837 & - & - \\
 & NIQE \textdownarrow & \textbf{4.012} & 4.653 & - & - \\
 & NIMA \textuparrow & \textbf{5.451} & 5.440 & - & - \\
\midrule
\multirow{5}{*}{}
 & Win rate \textuparrow & \textbf{76.83\%} & 23.17\% & - & - \\
 & Infer. time (2 obj) \textdownarrow & \textbf{5.13s} & 15s & - & - \\
 & Infer. time (8 obj) \textdownarrow & \textbf{8.15s} & 41.3s & - & - \\
 & Memory (2 obj) \textdownarrow & \textbf{5 GBs} & 7.19 GBs & - & - \\
 & Memory (8 obj) \textdownarrow & \textbf{5.5 GBs} & 7.19 GBs & - & - \\
\bottomrule
\end{tabular}
\caption{Quantitative comparison, inference cost, and user study win rate. Baselines: BAS (Build-a-Scene~\cite{eldesokey2024buildasceneinteractive3dlayout}), LooseCtrl (LooseControl~\cite{bhat2023loosecontrol}), LayoutG.\ (Layout-Guidance~\cite{ca_guidance}); ``obj'' denotes objects.}
\label{tab:metrics-results}
\vspace{-0.4cm}
\end{table}

\subsection{Dataset}
\label{sec:dataset}
LooseControl generates 3D annotations by lifting 2D segmentations to 3D. Since the official data and curation pipeline are not available, we developed our own pipeline to curate a dataset for fine-tuning our ControlNet.
% We leverage the dataset introduced by \cite{rigo2025esploraenhancedspatialprecision}, which consists of text descriptions, depth maps, and 2D bounding boxes paired with high-quality images generated by Flux.1 \cite{flux2024}.
We employ the dataset from \cite{rigo2025esploraenhancedspatialprecision}, containing 22k+ high-quality Flux.1 \cite{flux2024} images. These images are annotated with text, depth maps, and 2D bounding boxes, and are characterized by dense layouts containing multiple objects across 300 categories.

We lift the provided 2D boxes to 3D using their average depth and optimize per-scene camera poses by minimizing $1 - \text{mIoU}$ between the projected 3D boxes and the ground-truth 2D boxes; the recovered pose yields the paired 3D layout depth map used to fine-tune our ControlNet.

\subsection{Implementation Details}
We fine-tune ControlNet on 4,000 3D layout depth maps using a batch size of 2 and a learning rate of $1 \times 10^{-5}$ for 2.5 epochs.
% To facilitate comparison with prior work, we employ Stable Diffusion (SD) 1.5 \cite{sd} as our diffusion backbone. noting that our method is model-agnostic.
However, applying IP-Adapter directly to SD1.5 can yield suboptimal outputs, with authors recommending the use of community checkpoints \cite{ip_adapter_2023_github}.
% However, it has been observed that applying IP-Adapter directly to the base SD 1.5 model can yield suboptimal outputs, with authors recommending the use of fine-tuned community checkpoints \cite{ip_adapter_2023_github}.
Consequently, we adopt a publicly available SD1.5 checkpoint fine-tuned on high-quality Flux.1 outputs.
% Consequently, we adopt a community-made SD1.5 checkpoint fine-tuned on Flux.1 outputs, aimed at enhancing visual fidelity.

\subsection{Evaluation Protocol}
We adopt the evaluation protocol established by the baseline Build-a-Scene (BAS) \cite{eldesokey2024buildasceneinteractive3dlayout}.
We compare our method against BAS, LooseControl, and include Layout-Guidance \cite{ca_guidance} to provide a comparison with a 2D layout-guided method; for the latter, we derive 2D bounding boxes by projecting the 3D layout masks onto the image plane.
The evaluation is divided into two tasks: 3D layout control and consistency under layout change. BAS introduces two benchmarks, each comprising 100 scenes, with 5 random seeds each for fair comparison. The first benchmark consists of scenes with two random objects, where the objective is to generate an image that is consistent with the provided layout. The second benchmark focuses on editing, requiring a random object to be repositioned to a new location within the image.
Individual object descriptions are concatenated into a global prompt for global T2I alignment metrics.

\subsection{Evaluating 3D Layout Control}
\label{sec:generation-eval}
Given a layout defined by 3D bounding boxes and corresponding text prompts, the goal is to synthesize an image that adheres to these spatial and semantic constraints.

% Following BAS, we evaluate \ours{} on their proposed benchmark using the following metrics:
% \begin{itemize}
%     \item $\text{CLIP}_{\text{T2I}}$: To measure the semantic alignment between the output and the prompt, we compute the CLIP score \cite{radford2021learning} between the generated image and the global scene description $P_*$.
%     \item \textit{Object Accuracy (OA)}: To assess whether all requested objects are generated, we employ YOLOv8 \cite{reis2023real} to detect the presence of objects specified by the layout.
%     \item \textit{Mean Intersection-over-Union (mIoU)}: We compute the IoU between the bounding box predicted by YOLOv8 and the 2D bounding box projected from the 3D layout. This metric quantifies how accurately the object is positioned within the designated layout box.
% \end{itemize}

Following BAS, we evaluate \ours{} with three metrics.
$\text{CLIP}_{\text{T2I}}$ measures semantic alignment between the generated image and the global scene prompt via the CLIP~\cite{radford2021learning} score.
Object Accuracy (OA) uses YOLOv8~\cite{reis2023real} to check whether all requested objects are present.
Mean Intersection-over-Union (mIoU) measures spatial accuracy as the IoU between YOLOv8's predicted box and the 2D projection of the 3D layout box.

The top "Layout Control" part of \cref{tab:metrics-results} summarizes the quantitative results on the BAS 3D layout-to-image benchmark, averaged over 5 seeds. Our method, \ours{}, achieves the second-highest mIoU, closely approaching the performance of BAS, while achieving 78,3\% in object accuracy (OA), exceeding BAS by a substantial margin of 24.8\%, indicating a stronger ability to follow the intended layout. For $\text{CLIP}_{T2I}$, \ours{} outperforms all competing methods, demonstrating that its generated outputs align more faithfully with the textual prompts.

\begin{figure}[t]
    \centering
    \includegraphics[width=1\linewidth]{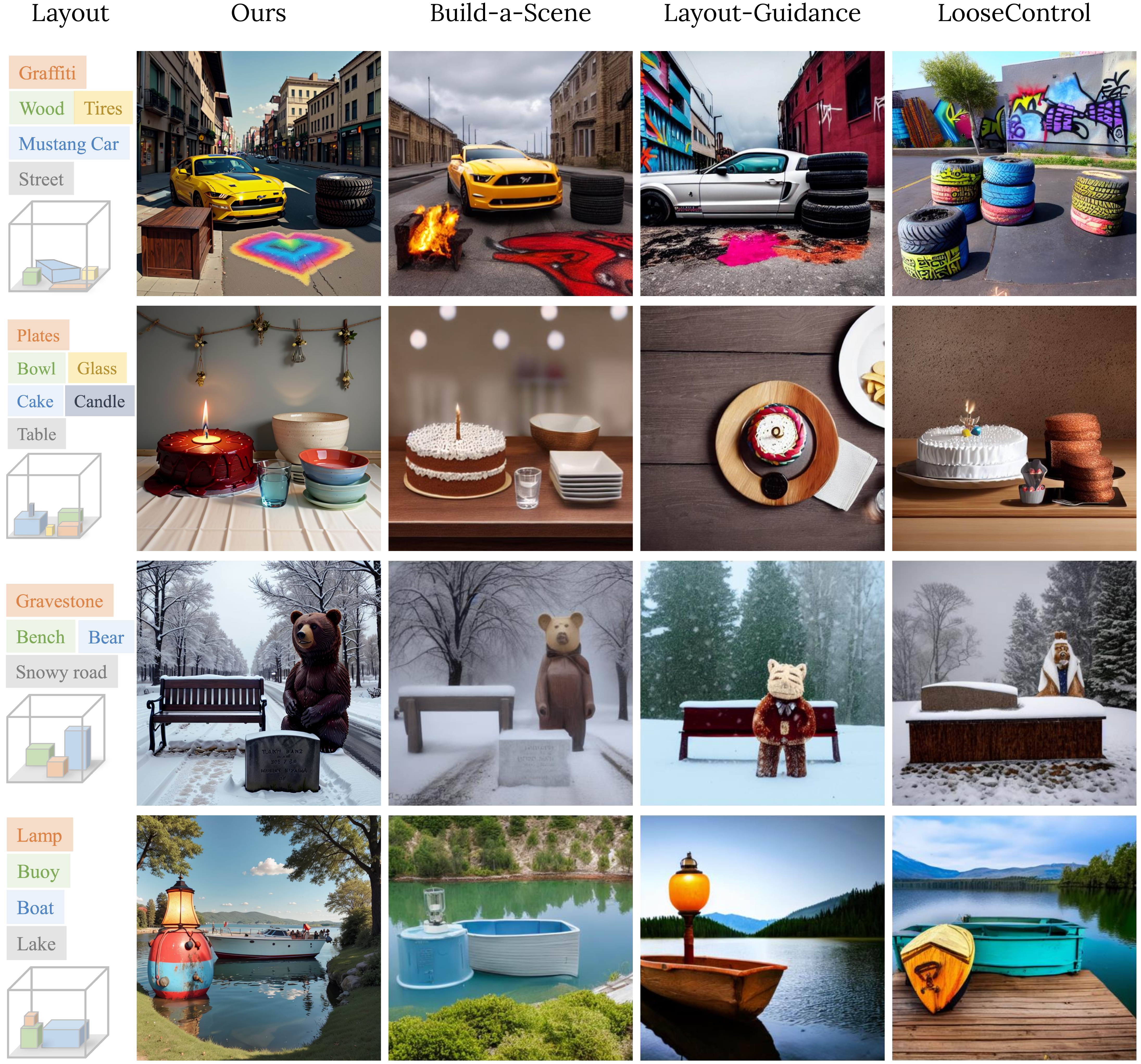}
    \caption{Qualitative comparison on the 3D layout control task.}
    \label{fig:layout-control}
    \vspace{-0.4cm}
\end{figure}

We present qualitative comparisons against all evaluated baselines in \cref{fig:layout-control}. As shown in the figure, our method achieves superior harmonization, layout fidelity, and overall image quality. LooseControl frequently fails to generate all specified objects, while Layout-Guidance struggles to accurately follow the provided layout. Both exhibit notably lower visual quality. Build-a-Scene can place objects in the correct locations, but its outputs lack realism: objects often appear box-like, poorly rendered, and insufficiently integrated into the surrounding scene.
In contrast, our method consistently produces visually coherent results, with objects that are realistic, correctly positioned, and well blended within the overall composition.

\begin{figure}[t]
    \centering
    \includegraphics[width=0.6\linewidth]{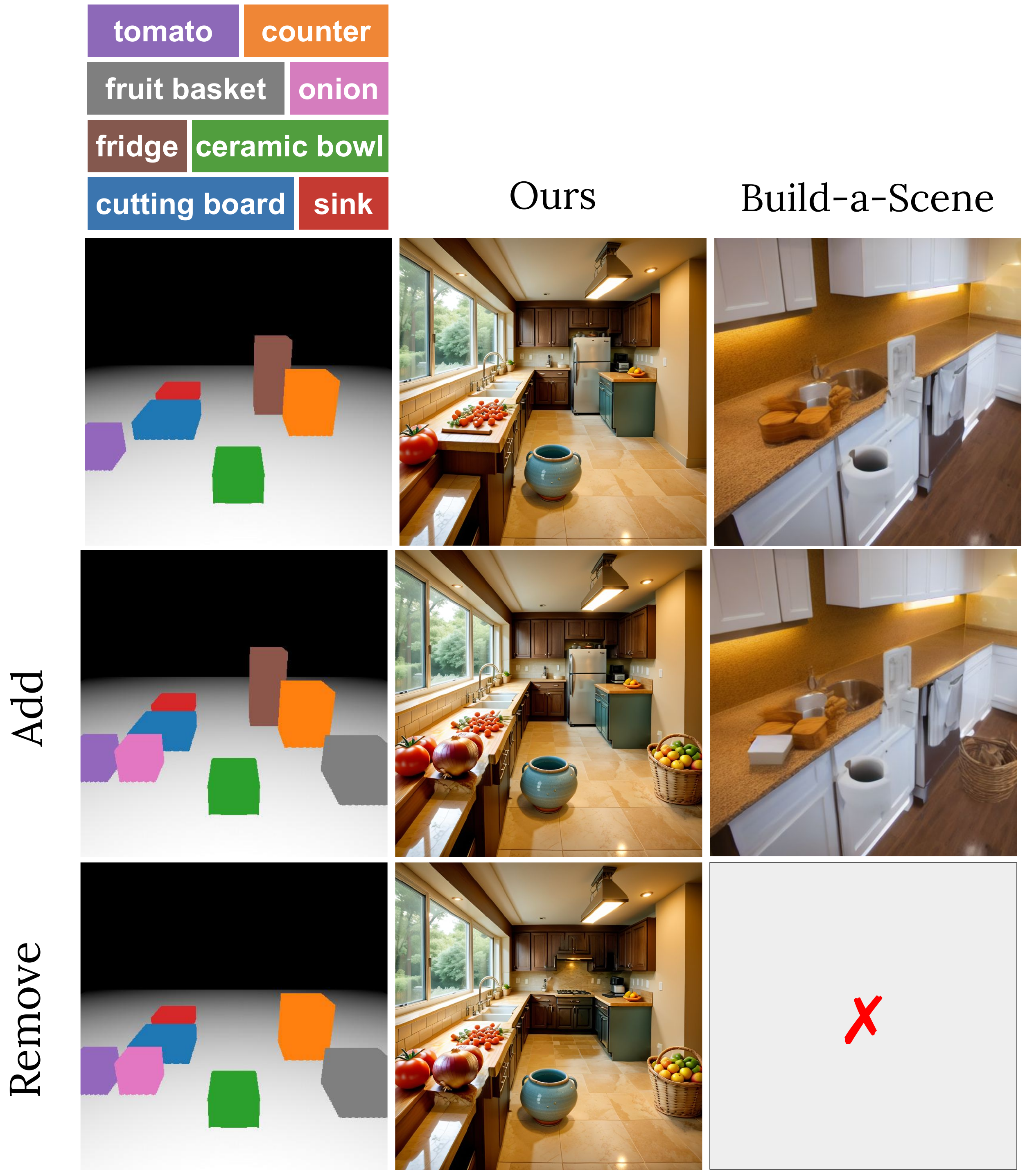}
    \caption{Qualitative comparison for adding and removing objects while keeping the scene consistent.}
    \label{fig:add-quals}
    \vspace{-0.4cm}
\end{figure}

\begin{figure}[t]
    \centering
    \includegraphics[width=0.9\linewidth]{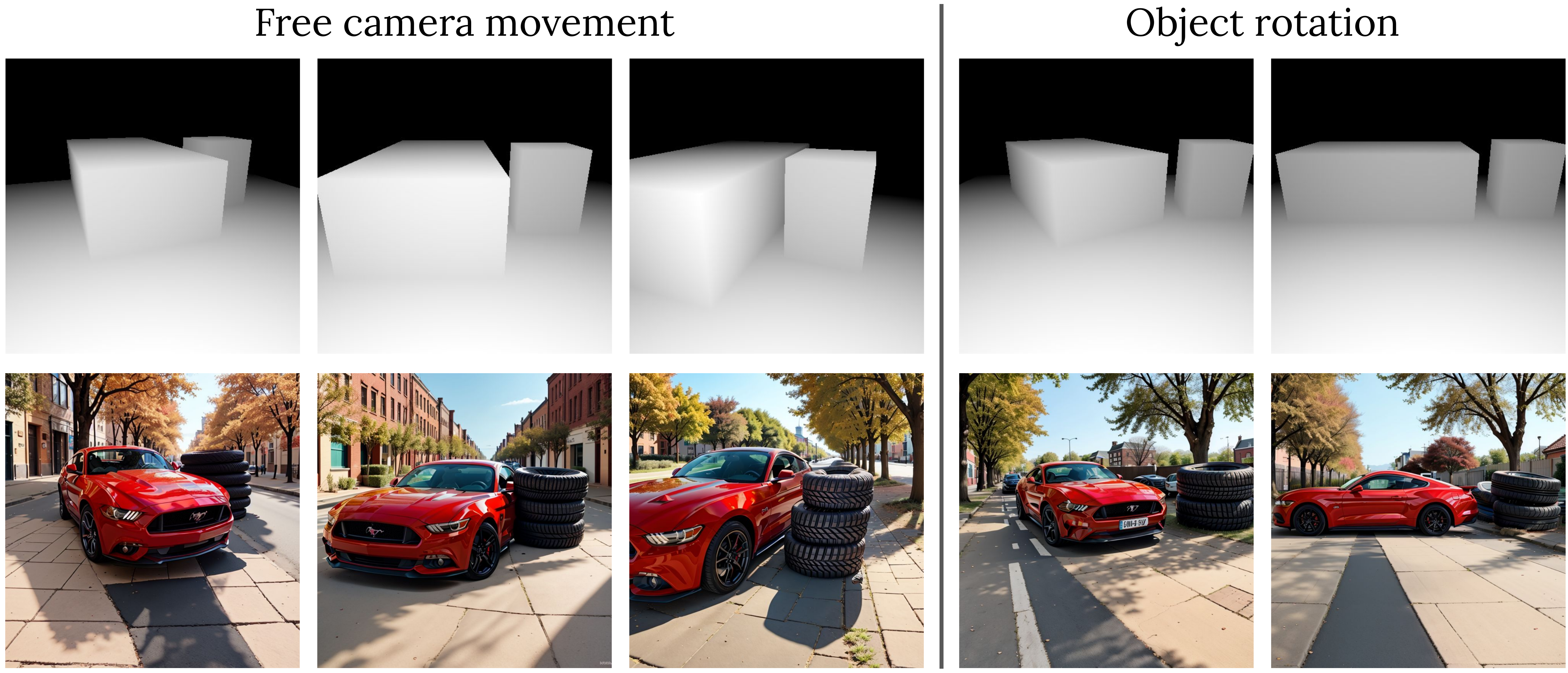}
    \caption{Examples of free camera control and object rotation that \ours{} can perform.}
    \label{fig:free-cam-rotation}
    \vspace{-0.4cm}
\end{figure}

We also evaluate \ours{} and BAS by rendering a complex scene with several objects, iteratively adding new objects, and removing one at the end.
% The results, along with the corresponding 3D layouts and text prompts, are shown in \cref{fig:add-quals}.
As shown in \cref{fig:add-quals}, BAS struggles to generate a coherent initial scene and fails to properly integrate added objects, resulting in unnatural appearances; since BAS cannot remove objects, we represent this action with a gray square. In contrast, \ours{} generates the whole starting scene in a single pass, harmonizing all objects jointly at each diffusion step, yielding a more cohesive result.
Furthermore, \ours{} unified representation allows for spatial manipulations, such as object rotation and free camera navigation, as shown in \cref{fig:free-cam-rotation}.
% In summary, \ours{} achieves efficient, high-fidelity synthesis while offering a versatile set of editing capabilities.

We further assess scalability by measuring inference time and peak GPU memory on scenes with 2 and 8 objects under identical hardware (\cref{tab:metrics-results}). BAS's iterative strategy yields constant memory but inference time that grows sharply with object count, whereas \ours{}'s parallel diffusion scales favorably on both axes.

\subsection{Evaluating Consistency under Layout Change}
\label{sec:editing-eval}
This task assesses the model's ability to preserve the visual identity of an object while undergoing translation or scaling, by measuring the visual similarity of the target object before and after the layout update.
% This task assesses the model's ability to preserve the visual identity of an object while undergoing spatial transformations, such as translation or scaling, within a provided layout. The evaluation quantifies the visual similarity of the target object before and after the layout modification.

Following BAS's protocol, we evaluate \ours{} by extracting crops of the target object from the original and edited images and computing two consistency metrics.
\textit{Crop $\text{CLIP}_{I2I}$} measures high-level semantic consistency via the CLIP score between pre-edit and post-edit crops.
\textit{Crop PSNR} measures low-level pixel fidelity between the original and transformed crops.
% Following the protocol established by BAS, we evaluate \ours{} on the corresponding benchmark by extracting crops of the target object from both the original and the edited images to compute consistency metrics. Specifically, we measure high-level semantic consistency using \textit{Crop $\text{CLIP}_{I2I}$}, which computes the CLIP score between the pre-edit and post-edit object crops. Additionally, we quantify low-level visual reconstruction quality via \textit{Crop PSNR}, which calculates the pixel-level fidelity between the original and the transformed object crops.
However, we argue that BAS's consistency metrics are inherently biased toward copy-paste approaches: by comparing isolated object crops, they favor methods like BAS that retain the original pixels, yielding high scores regardless of global scene quality or harmonization.
% This effectively bypasses the challenge of realistically integrating the object into a new context while producing high quality images.
% By comparing cropped objects in isolation, these metrics favor approaches like BAS, which achieves high consistency scores by simply retaining the original pixels, regardless of whether the resulting composition is photorealistic or semantically coherent.
% This effectively bypasses the challenge of realistically integrating the object into a new context while producing high quality images.

\begin{figure}[t]
    \centering
    \includegraphics[width=0.8\linewidth]{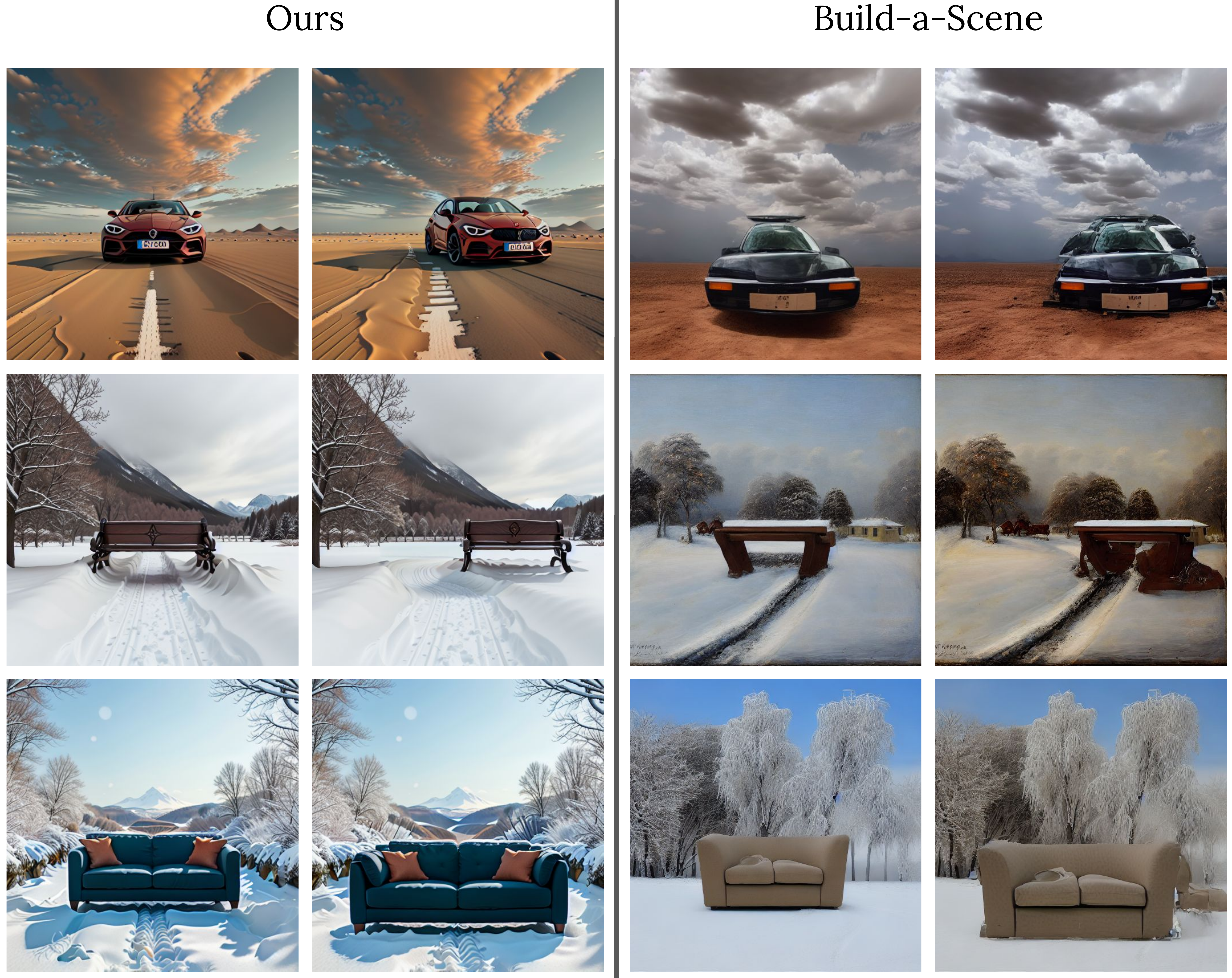}
    \caption{Qualitative comparison for consistency editing the scene under layout change.}
    \label{fig:edit-quals}
    \vspace{-0.4cm}
\end{figure}

As BAS leads the other baselines on crop-based metrics, we further compare \ours{} against BAS with a broader set of image-quality metrics: CLIP$_{\text{I2I}}$ and DINOv2~\cite{oquab2024dinov2learningrobustvisual} for global semantic and structural consistency; LPIPS~\cite{lpips} and DISTS~\cite{dists} for perceptual and textural fidelity; CLIP-IQA~\cite{clipiqa}, ImageReward~\cite{xu2023imagerewardlearningevaluatinghuman}, and NIMA~\cite{nima} for aesthetic appeal; and ARNIQA~\cite{agnolucci2023arniqalearningdistortionmanifold} and NIQE~\cite{Mittal2013MakingAniqe} for technical quality and naturalness.
We also conducted a user study with 50 participants who compared 30 random image pairs from our method and BAS, selecting their preference based on scene harmonization and visual quality.
\cref{tab:metrics-results} summarizes the quantitative results and reports the user study win rates.
% \cref{tab:metrics-results} summarizes the quantitative results on the BAS consistency benchmark and the win rate from the user study.
\ours{} closely rivals BAS on crop-based metrics, despite the metrics bias, while consistently outperforming all other methods in terms of image quality.
% \ours{} closely rivals BAS on crop-based metrics, despite the inherent advantage these metrics afford to copy-paste approaches, while consistently outperforming all other methods in terms of image quality.
In the user study, users preferred our method in 76.83\% of cases.

\Cref{fig:edit-quals} presents qualitative comparisons for 3D layout editing.
While BAS successfully relocates the target object, the generation quality remains low; objects often lack distinctive semantic features or suffer from severe geometric distortions due to the warping process. Furthermore, the edited objects are poorly integrated into the scene, lacking shadows and realistic boundary blending.
%lacking essential visual cues such as shadows and realistic boundary blending.
% It is also worth noting that BAS requires storing the background image, as it relies on extracting and re-pasting the object in a different location, and it is incapable of removing the object from the old location.
As demonstrated in \cref{fig:add-quals}, BAS also struggles with multi-object generation, often producing distorted or unrecognizable shapes, which renders its segmentation-based warping mechanism infeasible in complex scenes.
% As demonstrated in Figure \ref{fig:add-quals}, BAS struggles with multi-object generation, often producing distorted or unrecognizable shapes. Consequently, the segmentation-based warping mechanism employed by BAS, which relies on object extraction, is infeasible on such degraded outputs.
Instead, our method can handle both simple and complex scenes, seamlessly blending multiple objects in the scene and moving objects while maintaining visual consistency.

\subsection{Qualitative Demonstration on Surveillance Scenarios}
\label{sec:surveillance-qual}
\begin{figure}[t]
    \centering
    \includegraphics[width=\linewidth]{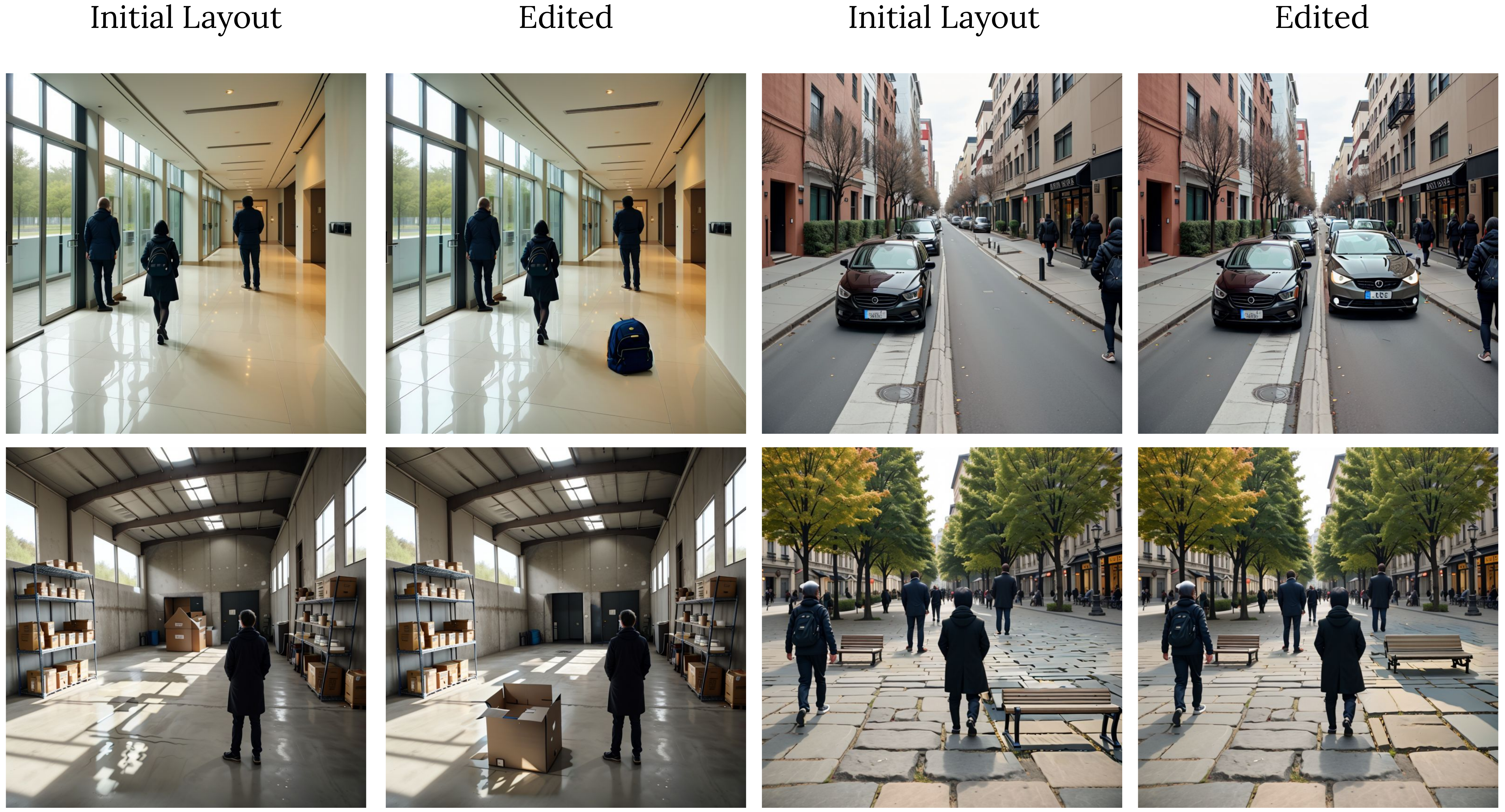}
    \caption{Qualitative results of \ours{} on four representative surveillance scenarios.}
    \label{fig:surveillance-quals}
    \vspace{-0.4cm}
\end{figure}

To demonstrate \ours{}'s applicability as a synthetic data generator for surveillance, in \cref{fig:surveillance-quals} we present qualitative results across four representative scenarios, each shown as a pair of images generated by our method from a fixed camera viewpoint: a base image generated from a 3D layout and an edited variant.
In a glass-walled indoor corridor, we insert and remove an unattended backpack, mirroring the abandoned-object setup used to train anomaly detection models.
In a warehouse scene, we move a cardboard box in front of the occupant, producing the controlled object-placement variation typical of activity-monitoring data.
In a residential street, we add a second car next to the one already parked, yielding the kind of vehicle-density variation useful for traffic-camera augmentation.
Finally, in a pedestrian plaza, we translate a bench between locations, illustrating spatial variation of static street furniture for diverse layout configurations.
Across all scenarios, \ours{} preserves the global scene composition, camera geometry, and lighting, modifying only the targeted region and supporting the creation of annotated training variations from a single base layout.

% \begin{table}[]
% \centering
% \begin{tabular}{lcccccccccc}
% \toprule
%  & ${CLIP}_{I2I}$ \textuparrow & DINO \textuparrow & LPIPS \textdownarrow & DISTS \textdownarrow & ARNIQA \textuparrow & CLIP-IQA \textuparrow & NIQE \textdownarrow & NIMA \textuparrow & OA \textuparrow & ImageReward \textuparrow \\ \midrule
% Build-a-Scene \cite{eldesokey2024buildasceneinteractive3dlayout} & 96.3978 & 0.8733 & 0.1538 & 0.1177 & 0.6703 & 0.8368 & 4.6532 & 5.4398 & \textbf{0.378} & 0.8627 \\
% \textbf{\ours} & \textbf{97.0683} & \textbf{0.9186} & \textbf{0.1153} & \textbf{0.0811} & \textbf{0.6731} & \textbf{0.8839} & \textbf{4.0115} & \textbf{5.4513} & 0.366 & \textbf{1.2808} \\ \bottomrule
% \end{tabular}
% \caption{aaaa}
% \label{tab:our-metrics-results}
% \end{table}

% \begin{table}[]
% \centering
% \begin{tabular}{@{}lccc|cc@{}}
% \toprule
%  & \multicolumn{3}{c|}{3D Layout Control} & \multicolumn{2}{c}{Object Consistency} \\
%  & $\text{CLIP}_{T2I}$ \textuparrow & OA \textuparrow & mIOU \textuparrow & Crop $\text{CLIP}_{T2I}$ \textuparrow & Crop PSNR \textuparrow \\ \midrule
% Layout-Guidance \cite{ca_guidance} & 0.323 & 48.2 & 0.425 & 0.838 & 28.35 \\
% LooseControl \cite{bhat2023loosecontrol} & 0.302 & 24.3 & 0.633 & \textbf{0.924} & 29.12 \\
% Build-a-Scene \cite{eldesokey2024buildasceneinteractive3dlayout} & 0.321 & 53.5 & \textbf{0.772} & \textbf{0.924} & \textbf{29.3} \\
% \textbf{\ours} & \textbf{0.3328} & \textbf{78.3} & 0.7278 & 0.919 & 29.10 \\ \bottomrule
% \end{tabular}
% \caption{Summary of quantitative results.}
% \end{table}

\section{Conclusion}
% === ORIGINAL CONCLUSION ===
% In this paper, we introduced \ours{}, a unified diffusion framework designed to address the limitations of existing text-to-image models in achieving precise, 3D-aware spatial control and consistent interactive editing. By integrating a fine-tuned depth-conditioned ControlNet with Blended Latent Diffusion, \ours{} enables the one-shot synthesis of complex scenes where individual objects are governed by specific 3D bounding boxes and distinct prompts. This approach effectively overcomes the ambiguity of global prompting and the computational inefficiencies of iterative generation pipelines found in prior works.
% Furthermore, we proposed a novel, warping-free editing pipeline that fundamentally improves object manipulation. Instead of relying on "copy-warp-paste" strategies that often degrade geometric fidelity, our method leverages IP-Adapter to regenerate objects at new locations while preserving their visual identity and seamlessly inpainting the background. Through extensive quantitative and qualitative experiments, we demonstrated that \ours{} significantly outperforms state-of-the-art baselines in 3D layout controllability, text-image alignment, and visual realism. By offering robust control over object placement, removal, and transformation without compromising scene coherence, \ours{} establishes a new standard for geometrically consistent and interactive 3D-guided image generation.
% === END ORIGINAL CONCLUSION ===
In this paper, we introduced \ours{}, a unified diffusion framework for generating images from 3D bounding-box layouts with fine-grained, per-object semantic control. By integrating a fine-tuned depth-conditioned ControlNet with Blended Latent Diffusion, \ours{} synthesises complex multi-object scenes in a single forward pass, binding individual text descriptions to specific 3D locations. This design overcomes the ambiguity of global prompting and the computational inefficiencies of iterative generation pipelines found in prior works.
Furthermore, we proposed a warping-free editing pipeline that supports interactive object insertion, removal, and spatial transformation via regeneration rather than pixel deformation, leveraging IP-Adapter to preserve object identity across edits. This capability enables efficient creation of diverse scene variations from a single base layout, a property particularly valuable for surveillance data augmentation, where generating large-scale, annotated datasets with controlled spatial configurations is essential yet collecting real data remains costly and privacy-sensitive.
Through extensive quantitative and qualitative experiments, we demonstrated that \ours{} significantly outperforms state-of-the-art baselines in 3D layout controllability, exceeding BAS by 24.8\% in object accuracy, as well as in text-image alignment and visual realism, with users preferring our method in 76.83\% of a 50-participant study, while generating 8-object scenes roughly 5$\times$ faster than BAS. This establishes \ours{} as a controllable synthetic data generator with properties directly suited to surveillance data needs: fixed-viewpoint consistency, per-object annotation, privacy compliance, and efficient generation of diverse scene variations.

% \textbf{Limitations.} 3D bounding boxes approximate roughly convex objects well but fit irregular shapes (trees, articulated poses) more coarsely, and heavy box intersections can produce occlusion artifacts; finer-grained representations (meshes, convex hulls) are a natural extension. Because the depth map encodes only camera-visible surfaces, transformations that reveal a new face (e.g., 180-degree rotation) depend on the model's learned prior rather than the layout, an explicit trade-off for the lightweight bounding-box interface, which richer representations such as CustomNet's~\cite{yuan2023customnet} transformation matrices could address. Finally, our ControlNet is fine-tuned on Flux.1-generated images and we have not yet quantitatively measured downstream-task impact on real surveillance benchmarks~\cite{farley2023diffusion_ped_det,seib2024synth_ped_gan}; broader real-world validation remains future work.

\bibliographystyle{IEEEtran}
\bibliography{main}

\clearpage
\appendix
\section{Ablations}
We ablate the main design choices of the pipeline, including the use of vanilla versus fine-tuned SD1.5, the introduction of a background-preserving mask, different ControlNet adaptation strategies, and the choice between synthetic and real training data. Experiments are conducted on a subset of the BAS consistency benchmark and a subset of the metrics in \cref{sec:editing-eval}. As shown in \cref{tab:ablations}, introducing the background mask consistently improves most metrics, confirming its importance for preserving scene regions and producing more stable edits. Using a community fine-tuned SD1.5 model further improves visual quality on several metrics.

We additionally compare full fine-tuning of the ControlNet with a LoRA-based adaptation following LooseControl. While the LoRA-based approach is slightly better on some consistency metrics, it yields lower image quality scores. Since our ControlNet is primarily trained on synthetic data, raising potential concerns about domain bias, we also fine-tune it on real images from the MS-COCO dataset. This variant shows slight improvements on some metrics but similarly exhibits lower image quality. Overall, we opt for a fully fine-tuned ControlNet trained on synthetic data, as it provides the best trade-off and achieves superior image quality compared to the other variants.

\begin{table}[h!]
\scriptsize
\centering
\setlength{\tabcolsep}{3pt}
\begin{tabular}{@{}c|lcccccc@{}}
\toprule
 & & \multicolumn{6}{c}{\ours{} variants} \\
\midrule
\multirow{4}{*}{\rotatebox{90}{\makecell{Design\\Choices}}}
& Data &
Synthetic & Synthetic & Synthetic & Synthetic & Real & Real \\
& ControlNet tuning &
Full & Full & Full & LoRA & Full & Full \\
& Fine-tuned SD1.5 & 
\xmark & \xmark & \checkmark & \checkmark & \checkmark & \xmark \\
& Background mask &
\xmark & \checkmark & \checkmark & \checkmark & \checkmark & \checkmark \\
\midrule
\multirow{6}{*}{\rotatebox{90}{\makecell{Metrics}}}
& $\text{CLIP}_{I2I}\uparrow$ &
96.955 & 97.068 & 96.987 & \textbf{97.770} & 97.694 & 97.235 \\
& DINOv2 $\uparrow$ &
0.906 & 0.919 & 0.895 & \textbf{0.938} & 0.917 & 0.919 \\
& DIST $\downarrow$ &
0.099 & 0.081 & 0.082 & \textbf{0.061} & 0.084 & 0.078 \\
& NIQE $\downarrow$ &
4.799 & 4.012 & 4.009 & 3.980 & 4.097 & \textbf{3.636} \\
& NIMA $\uparrow$ &
5.479 & 5.451 & \textbf{5.625} & 5.350 & 5.302 & 4.951 \\
& ImageReward $\uparrow$ &
1.278 & \textbf{1.281} & 1.265 & 0.959 & 1.229 & 0.933 \\
\bottomrule
\end{tabular}
\caption{Ablation study.}
\label{tab:ablations}
\end{table}

Our depth-conditioned ControlNet is primarily trained on synthetic Flux.1 data, which raises potential concerns regarding domain shift when applied to real-world images. We partially explore this issue by fine-tuning the model on real MS-COCO images; however, this experiment yields only limited improvements on some consistency metrics and often results in reduced perceptual image quality. These results indicate that the simple introduction of real data does not necessarily lead to better robustness or generalization in our setting. We attribute this behavior to discrepancies between synthetic and real data distributions, including differences in depth quality, noise, and scene composition. While a more comprehensive evaluation on real data would be required to draw stronger conclusions, such an investigation is beyond the scope of this work. Overall, our findings suggest that synthetic training provides a reasonable balance between consistency and visual quality, but further study is needed to better understand and mitigate domain shift effects.

\section{Time and Memory Scalability}
\begin{figure}
    \centering
    \includegraphics[width=\linewidth]{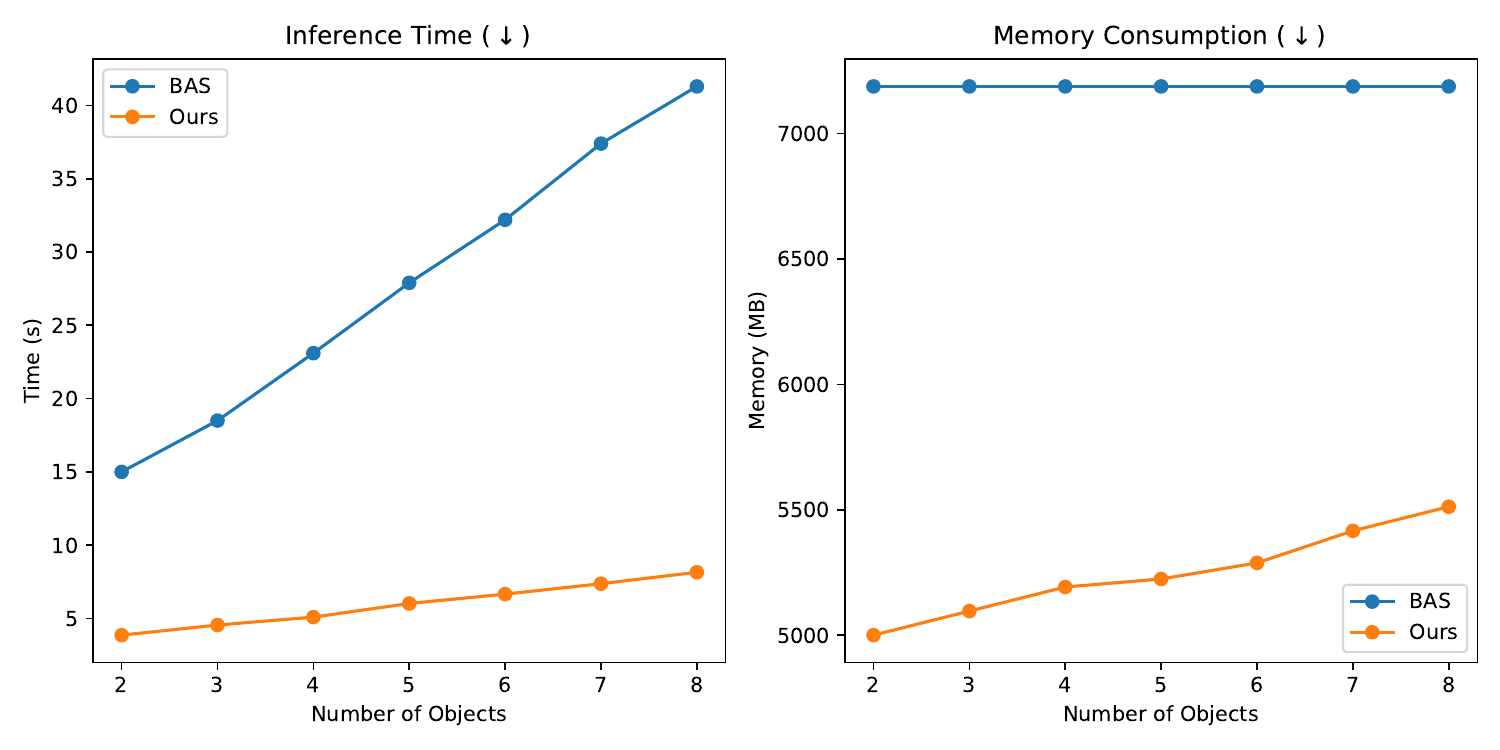}
    \caption{Inference time and memory comparison with increasingly crowded scenes for the 3D layout control task.}
    \label{fig:generation-time-mem}
\end{figure}

To evaluate the scalability of our approach in increasingly dense scenes, we generate synthetic scenes containing between 2 and 8 objects plus background, and measure both inference time and peak GPU memory consumption. We perform the same evaluation for BAS under identical hardware conditions and report the results in \cref{fig:generation-time-mem}. Due to its iterative generation strategy, BAS exhibits constant memory usage across different scene complexities, but its inference time increases approximately linearly with the number of objects. In contrast, \ours{} generates the entire scene in a one go by running multiple diffusion processes in parallel. As a result, memory consumption increases moderately with the number of objects, while remaining substantially lower than that of BAS. Moreover, the parallel nature of \ours{} allows inference time to scale much more favorably, exhibiting only a marginal increase as scene complexity grows.

\section{User study protocol}
We provide more details about the user study to ensure reproducibility.
We conducted a controlled user study with 50 participants. Each participant was shown 30 image pairs, randomly sampled from the test set, where each pair consisted of the result produced by our method and by BAS on the same composite input. The left–right ordering of the two methods was randomized independently for each comparison to avoid positional bias. Participants were asked to select the image they preferred based on overall scene harmonization and visual quality, without being informed of which method produced each result. No time limit was imposed. In total, the study collected 1,500 pairwise comparisons, from which we report the win rate of each method.

\end{document}